\documentclass[10pt, a4paper]{article}

\usepackage[final]{lrec2026} 

\title{Generating High Quality Synthetic Data for Dutch Medical Conversations}

\name{Cecilia Kuan, Aditya Kamlesh Parikh, Henk van den Heuvel} 

\address{Center for Language Studies, Radboud University\\
         Nijmegen, Netherlands\\
         \{cecilia.kuan, aditya.parikh, henk.vandenheuvel\}@ru.nl\\}

\abstract{
Medical conversations offer insights into clinical communication often absent from Electronic Health Records. However, developing reliable clinical Natural Language Processing (NLP) models is hampered by the scarcity of domain-specific datasets, as clinical data are typically inaccessible due to privacy and ethical constraints. To address these challenges, we present a pipeline for generating synthetic Dutch medical dialogues using a Dutch fine-tuned Large Language Model, with real medical conversations serving as linguistic and structural reference. The generated dialogues were evaluated through quantitative metrics and qualitative review by native speakers and medical practitioners. Quantitative analysis revealed strong lexical variety and overly regular turn-taking, suggesting scripted rather than natural conversation flow. Qualitative review produced slightly below-average scores, with raters noting issues in domain specificity and natural expression. The limited correlation between quantitative and qualitative results highlights that numerical metrics alone cannot fully capture linguistic quality. Our findings demonstrate that generating synthetic Dutch medical dialogues is feasible but requires domain knowledge and carefully structured prompting to balance naturalness and structure in conversation. This work provides a foundation for expanding Dutch clinical NLP resources through ethically generated synthetic data.
 \\ \newline  }

\begin{document}
\maketitleabstract

\section{Introduction} \label{sec:intro}


Recent developments in Natural Language Processing (NLP) have greatly advanced text analysis, especially in the medical domain. Specifically, analyzing physician-patient conversations through clinical NLP can enrich research datasets and provide data-driven insights into patient-initiated concerns, which are often absent from Electronic Health Records (EHRs) \citep{weiner2024accuracy,laleye2020french,alshaikhdeeb2025generation}. Such conversational data have also been shown to capture interactional detail and patient narratives valuable for prediction and research \citep{pyne2023analysis}, while potentially reducing clinician workload \citep{lukac2025randomized,zhang2024annotate}. 

The performance of NLP models strongly depends on the size, quality, and domain alignment of their training data. However, access to healthcare datasets is restricted by privacy regulations such as the General Data Protection Regulation\footnote{\url{https://gdpr-info.eu/}} (GDPR) \citep{marino2025medical,gassenn2025medical}, and anonymization processes can still risk potential re-identification \citep{el2009evaluating}. 

\begin{figure}[!ht]
\centering
\begin{center}
\includegraphics[width=0.9\columnwidth]{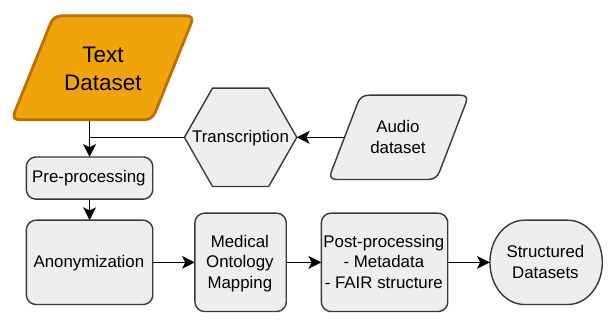}
\caption{Broader Project Workflow - this study focuses on the Text Dataset component (highlighted)}
\label{broader_workflow}
\end{center}
\end{figure}

Our broader objective is to convert unstructured Dutch medical data, both audio and text, into structured resources that enable research, analysis, and interoperability. As shown in Figure \ref{broader_workflow}, it involves multiple processing steps, including transcription, anonymization, medical entity recognition, and ontology mapping. 

Currently, our experiments rely on a small subset of audio–transcription pairs for controlled testing, which limits large-scale fine-tuning or comprehensive evaluation. This constraint reflects a broader challenge faced by the clinical NLP community: the scarcity of publicly available sensitive datasets \citep{hiebel2023can}. 

\begin{figure}[!ht]
\centering
\begin{center}
\includegraphics[width=\columnwidth]{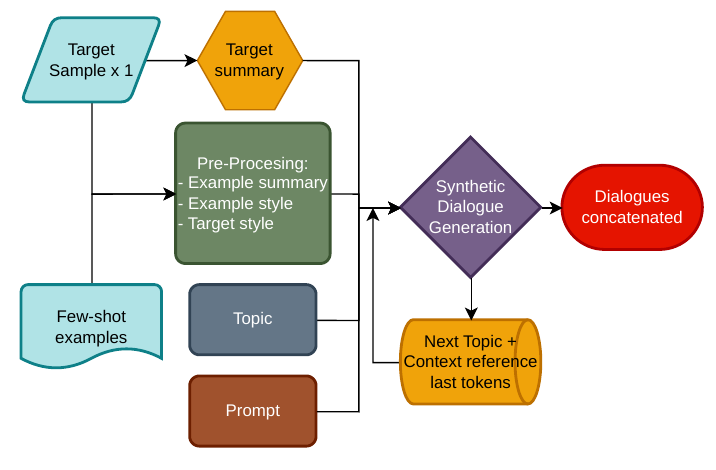}
\caption{Synthetic Dialogue Text Generation Workflow}
\label{fig:syn_txt_workflow}
\end{center}
\end{figure}

Figure \ref{fig:syn_txt_workflow} illustrates the workflow of this study, which is motivated by the broader objective and focuses specifically on generating synthetic Dutch medical text dialogues as a resource to support and evaluate the pipeline. To mitigate data scarcity, generating high-quality synthetic data offers a practical alternative by enabling model fine-tuning, benchmarking, and system validation without the privacy constraints of real clinical data. Synthetic data can therefore serve as a privacy-compliant substitute for real datasets, supporting broader research collaboration and performance benchmarking in Dutch clinical NLP \citep{ive2020generation}.

This paper presents a pipeline for generating synthetic Dutch medical text dialogues using a Large Language Model (LLM) fine-tuned with a Dutch dataset\footnote{Generated corpus available at: \url{https://doi.org/10.34973/mvpm-9987}} by leveraging real clinical conversations as a linguistic and structural reference with two-shot samples for guidance. Our key research question is:
\begin{quote}
    To what extent can a Dutch LLM produce synthetic medical dialogues that match real-world data for supporting clinical NLP pipeline development?
\end{quote}

To the best of our knowledge, no prior work has explored synthetic Dutch data generation using LLMs beyond clinical reports and EHR. Existing work on other languages and domains is reviewed in Section \ref{sec:related_work}. The proposed pipeline and evaluation provide a novel framework for generating and assessing synthetic data that can advance Dutch clinical research and NLP development. By combining quantitative and qualitative evaluation, this work also explores the limits of automatic metrics in measuring linguistic quality, emphasizing the importance of human review in assessing complex linguistic patterns in natural conversations.

\section{Related Work}\label{sec:related_work}


One major challenge in clinical NLP is data scarcity and privacy concerns surrounding patient information. Synthetic medical data generation provides a valuable alternative under these constraints. This approach has been explored for synthetic medical dialogues and clinical notes in several studies \citep {das2024synthetic, mianroodi2025medsynth}. In addition, models developed using synthetic data have shown performance comparable to or exceeding models developed by real-world data \citep{mianroodi2025medsynth,ive2020generation}.

Previous work on synthetic text generation include synthetic medical data generation in English \citep{meoni2024generating,melamud2019towards,ive2020generation,naguib2024few}, medical dialogue generation in English and other languages except Dutch \citep{hiebel2023can,wang2023umass_bionlp,lozoya2024generating,almutairi2024synthetic,mousavi2021would}, synthetic EHR generation in Dutch \citep{libbi2021generating}, and automatic medical reporting (summary) in Dutch \citep{van2023enhancing}. This review confirms that synthetic Dutch medical dialogue generation remains unexplored.

\section{Methodology} \label{sec:methodology}
In this section, we describe the experimental setup of synthetic Dutch medical dialogue generation in detail. Supporting materials are provided in the Appendices.

\subsection{Data}
In this work, we use a real-life dataset containing transcriptions of patient-doctor conversations in the nephrology domain from Nivel Institute\footnote{\url{https://nictiz.nl/}}'s archive collection as target samples for the LLM to adapt linguistic features for text generation. The nephrology focus reflects the scope of the broader project, which targets this clinical specialty. The manual transcriptions were created as part of the HoMed (Homo Medicinalis) project\footnote{\url{https://homed.ruhosting.nl/publications}}. Due to data protection constraints, only the textual transcriptions of the original audio recordings are accessible for research use.

The selected dataset originally contained nine transcriptions of nephrology consultations. Two files were used as few-shot examples, and the remaining seven were used as source material for synthetic dialogue generation. Several of these seven files were chunked into 1,000-token segments to accommodate the LLM's context window size, resulting in a total of nine usable text files. Each synthetic dialogue was modeled on the structure and linguistic style of its corresponding real dialogue, maintaining a one-to-one mapping between the reference and generated dialogues. File sizes range from 1,000 to 4,000 words and 170 to 650 speaker turns. The mean word count across target sample files is 2,708, and the mean number of turns is 396. 

\subsection{Model}
Meta’s Llama-3-8B-Instruct \citep{dubey2024llama}\footnote{\url{https://huggingface.co/meta-llama/Meta-Llama-3-8B-Instruct}} was initially chosen for its open-source accessibility and suitability for secure, privacy-preserving experiments. The smaller model was used due to computing constraints. While early outputs showed some inconsistency and repetition, these observations motivated further adaptation toward domain-specific fine-tuning for improved dialogue quality.

To explore alternatives, we selected Llama-3-ChocoLlama-8B-Instruct \citep{meeus2024chocollama}\footnote{\url{https://huggingface.co/ChocoLlama/Llama-3-ChocoLlama-8B-instruct}} (ChocoLlama in this paper), an instruction-tuned model fine-tuned on Dutch-translated instruction datasets. This model also offers open-source accessibility and local deployment, addressing privacy constraints. According to testing results reported by \citet{meeus2024chocollama}, it is the best performing model in the ChocoLlama family. The base model, Llama-3-ChocoLlama-8B-Base, was adapted from Meta's Llama-3-8B and fine-tuned on an extensive corpus of native Dutch text totaling approximately 32 billion tokens. As an instruction-tuned model, ChocoLlama aims to improve instruction following and the coherence of Dutch outputs, helping to address the limited Dutch coverage in general multilingual LLMs. The model is released under the CC BY-NC 4.0 license, which permits non-commercial use only.

\subsection{Experiment Setup}
Figure \ref{fig:syn_txt_workflow} shows the workflow for synthetic Dutch medical dialogue generation, which involves preparing examples and controlled information to guide the LLM in generating dialogues that are contextually relevant and linguistically coherent. Major steps are as follows:
\begin{itemize}
    \item Target sample summary generation and preprocessing - target sample summaries were generated as a reference for linguistic style and overall dialogue structure. Summaries reduced token consumption within the model's context window limit of approximately 8,000 tokens, leaving space for the prompt, few-shot examples, style specifications, and sliding window context. Bullet-point summaries also provided more structured guidance than raw dialogue text. Summaries were reviewed manually to ensure quality before using them for dialogue generation. 
    \item Few-shot learning example preprocessing - two dialogue files were used for few-shot learning. For each file, an initial segment (approximately 400 tokens) was summarized as input, while a later, non-overlapping segment (approximately 1,200 tokens) served as output, with a 100-token gap to avoid overlap between segments. Input-output pairs were designed to demonstrate stylistic features - turn structure, tone, sentence length - rather than content coherence.
    \item Other controlled information - medical specialty domain (nephrology in this study) and four topics (\textit{symptomen} / symptoms, \textit{medicatiegebruik} / medical use, \textit{leefstijl} / lifestyle, \textit{laboratoriumuitslagen} / laboratory results) are provided for dialogue generation
    \item Prompt and above information are collected and provided to LLM for text generation. Prompt information is provided in Section \ref{subsec:prompt}.
    \item Dialogue generation - One dialogue will be generated with each topic along with the other information. In order to maintain contextual continuity, the last 150 words of the last generated dialogue will be passed onto the LLM as part of the instruction for next dialogue generation.
    \item Generated Dialogue concatenation - dialogues generated using given topics will be concatenated as the final dialogue.
\end{itemize}

\subsection{Prompt Engineering} \label{subsec:prompt}
To identify an effective prompt, we first referenced the dialogue-generation prompt from \citet{mianroodi2025medsynth} to draft an initial version. Prompt engineering typically requires several iterations and adjustments to reach the desired output quality \citep{zhou2022large}. Following the approach of \citet{tang2023does}, we instructed the LLM to generate four variations and conducted a small-scale test to select the most suitable version. The selected version was then refined manually until it produced the intended responses.

The final prompt defined the LLM's role, speaker roles, the subject of the dialogue (medical domain, e.g., nephrology or oncology), a predefined list of conversational topics, limits on turn length, approximate total number of turns and words, a single sentence per turn (to mimic the target sample style), and integration of medical terminology. The prompt was designed to encourage professional yet natural dialogue suitable for clinical contexts. All prompts were written in Dutch. The complete prompt, which strongly influences dialogue quality and fluency, is provided in the Appendix \ref{app:prompt}.

\subsection{Evaluation Methods}
To assess the quality of synthetically generated dialogues, both quantitative and qualitative evaluations were conducted. Quantitative analysis assessed dialogue-level characteristics of fluency and realism, including turn alternation, greeting and closing phrases repetition, role consistency, Average Sentence Length (ASL), average Sentences Per Turn (SPT), topic coverage, and lexical diversity. Qualitative analysis focused on the practicality and clinical usability of the generated dialogues through human evaluation. 

\vspace{1em}
\noindent\textbf{Quantitative Analysis}

Turn alternation rate and the repetition of greeting or closing phrases were used to identify potential structural errors and assess dialogue organization and fluency--important for understanding downstream usability. Alternation rate was calculated as the proportion of speak switches within each dialogue. A perfect rate of 1.0 indicates strict alternation, which is rare in real conversations where short interjections (e.g.,"yes", "no") and overlaps commonly occur \citep{clark2002using,shriberg2001errrr}.

Role consistency evaluated how well speaker roles were represented through their lexical choices, since physician-patient communication show systematic vocabulary differences between the two roles. Prior studies have found that physicians typically use more technical and domain-specific terminology, whereas patients more often describe symptoms, experiences, or emotional state \citep{whittaker2024physicians,schillinger2021precision,pires2014communication}. This distinct lexical difference formed the basis for evaluating role consistency through keyword matching.

Keyword matching was performed between the generated dialogues and predefined role-specific lexicons. Relevant vocabularies were extracted from the Dutch medical ontology, \citet{SNOMED-NL}, using procedure terms for physicians and clinical finding or symptom terms for patients. The most frequent items in nephrology conversation transcripts (unseen during text generation) were identified from these vocabularies, and the top 300 items were selected for each role. As no prior work has evaluated role consistency in synthetic Dutch medical dialogues, no direct benchmarks are available. Role consistency scores were interpreted relative to a heuristic range of 0.05–0.35. Keyword lists for doctors and patients are provided in the Appendix \ref{app:keyword-lists}.

The ASL of the target sample was provided to the LLM as a reference, and prompts instructed one sentence per turn. ASL and SPT were measured to assess compliance with these structural instructions. Sentence length was defined as the number of words per sentence, ASL as the mean of all sentence lengths, and SPT as the mean number of sentences per turn.

Keyword-based evaluation, a common metric in Natural Language Generation, was adapted to assess topic generation since topics were explicitly defined in the prompt \citep{sun-etal-2023-evaluating}. Representative keywords were compiled for each topic, and their occurrences in the generated dialogues served as an evaluation metric. For medical topics such as \textit{symptomen} (symptoms), \textit{medicatiegebruik} (medication use), \textit{laboratoriumuitslagen} (laboratory results), keywords were extracted from clinically validated Dutch SNOMED CT ontology by \citet{SNOMED-NL}\footnote{NL release 20250930, https://nictiz.nl/wat-we-doen/activiteiten/terminologie/snomed/, last visited: October 2025}. For the non-medical topic "\textit{leefstijl}" (lifestyle), keywords were derived using semantic relations from the Open Dutch WordNet \citep{postma2016open} by \citet{ODWN}. The compiled keyword lists are provided in the Appendix \ref{app:keyword-lists}.

Lexical diversity measures linguistic richness, where excessive word repetition indicates reduced variety in natural conversations. The Type-Token Ratio (TTR), defined as the ratio of unique words to total words in a text \citep{rosillo2025entropy}, was used as a baseline measure for lexical diversity. However, it is known to decline as repetition increases naturally in longer text \citep{bestgen2024back}. To address this limitation, the Mean Segmental Type–Token Ratio (MSTTR) was also computed, estimating average TTR across fixed-size windows within a text \citep{rasanen2025pipeline}. A window length of 50 words was used to quantify lexical diversity in the generated medical dialogues. Given the mean dialogue length of 2,708 words, lower overall TTR values were expected, whereas specialized medical terminology may lead to relatively higher MSTTR scores.

\vspace{1em}
\noindent\textbf{Qualitative Analysis}

Human evaluation assessed contextual richness, natural language use, lexical appropriateness, and clinical relevance--aspects that are difficult to quantify through automated metrics \citep{tam2024framework}. Such evaluations are particularly valuable when no ground-truth or reference data are available for validation. A scoring rubric was used to standardize ratings across evaluators. The rubric, adapted from \citet{fraile2025expert} and customized to evaluate synthetic medical dialogues in this study, included five categories: coherence, consistency, fluency, relevancy, and clinical use. Each was rated on a five-point scale, where 0 indicates no adherence and 5 indicates full adherence to the category criteria. The complete rubric is provided in the Appendix \ref{app:rubric}.

Five native Dutch-speaking reviewers participated in the evaluation, four of whom were medical practitioners. The non-medical reviewer did not score the clinical use category. For each aspect, mean and standard deviation scores were computed, and Inter-Rater Reliability (IRR) was assessed using Krippendorff’s $\alpha$\citep{krippendorff2011computing}, which accommodates multiple raters and data types. Interpretation followed established guidelines \citep{Krippendorff2018ContentAnalysis, marzi2024k}, where $\alpha < 0$ indicates poor agreement, 0.00–0.20 slight, and 0.21–0.40 fair agreement.

Finally, correlations between quantitative and qualitative scores were examined using Spearman’s rank correlation coefficient ($\rho$). Positive $\rho$ values indicate alignment between automatic and human scores, whereas negative values denote inverse relationships.
\section{Results and Discussion}
\subsection{Quantitative Results}

\begin{figure}[!ht]
\begin{center}
\includegraphics[width=\columnwidth]{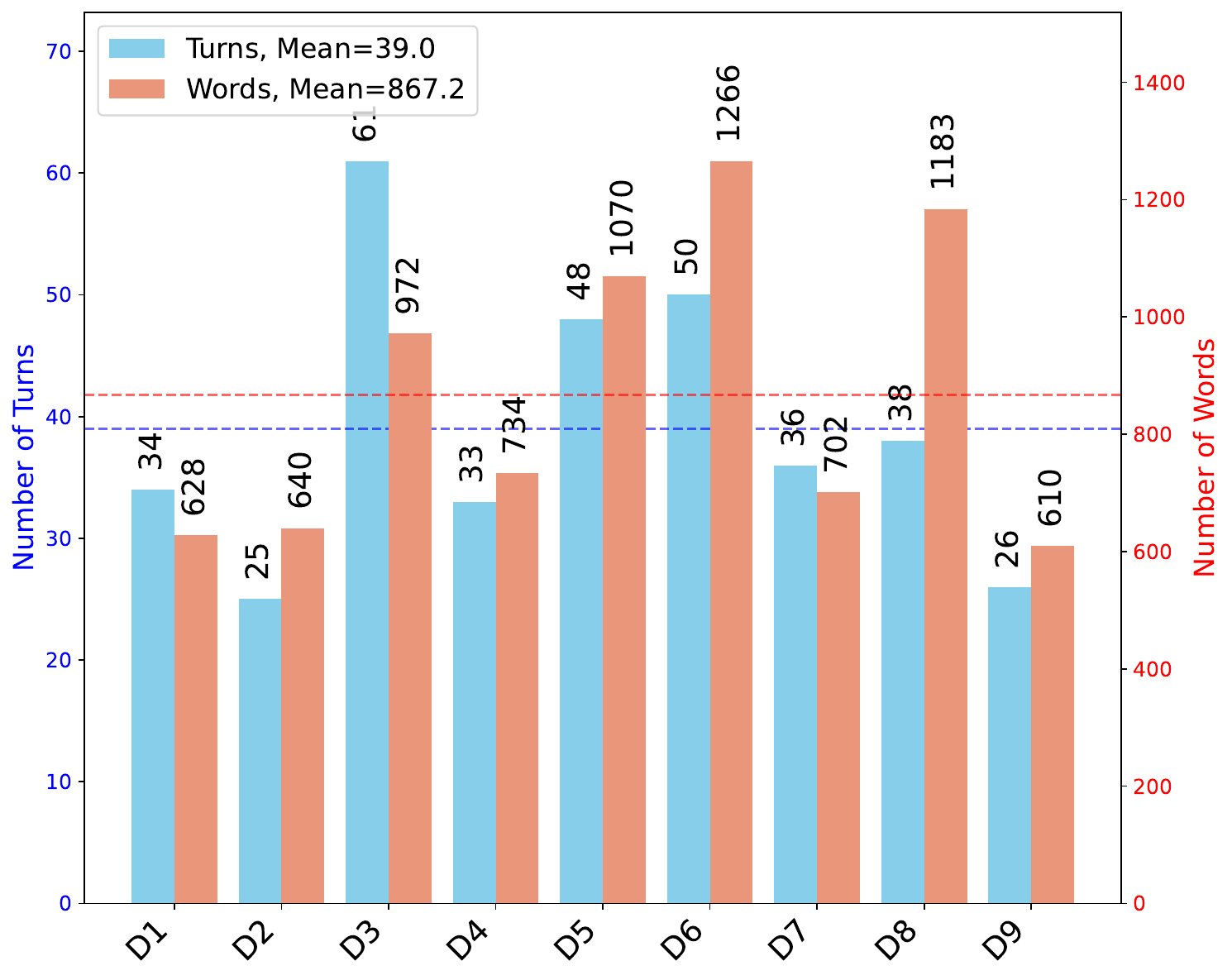}
\caption{Synthetic Dialogue Statistics - Number of Words and Turns. D1–D9 denote individual-generated dialogues.}
\label{fig:text_count_bar}
\end{center}
\end{figure}

We evaluated the generated dialogues using structural, lexical, and topic-based metrics. Figure~\ref{fig:text_count_bar} summarizes the distribution of word and turn counts across all generated dialogues.

The summary of results is presented in Table \ref{tab:evaluation_summary}.
\begin{table}[!ht]
\begin{center}
\footnotesize
\begin{tabularx}{\columnwidth}{lcc}
\textbf{Metric} & \textbf{Mean} & \textbf{SD} \\
\hline
\hline
Alternation rate & 0.973 & 0.021 \\
Role consistency & 0.012 & 0.007 \\
ASL & 16.18 & 4.03 \\
Average SPT & 2.14 & 0.37 \\
Topic coverage & 0.889 & 0.132 \\
TTR & 0.377 & 0.04 \\
MSTTR & 0.834 & 0.01 \\
&&\\
\textbf{Metric}& \textbf{Total Qty} &\\
\hline
\hline
Greeting Detection & 21 & - \\
Closing Detection & 9 & - \\
\end{tabularx}
\caption{Quantitative Evaluation Summary}
\label{tab:evaluation_summary}
\end{center}
\end{table}

\textbf{Alternation Rate.}
The mean alternation rate across the nine generated dialogues is 0.973. The near-perfect score suggests that the generated dialogues may be overly structured and show scripted turn-taking behavior rather than a real-life conversational flow. 

\textbf{Greeting/Closing Detection.} 
A total of 21 greeting occurrences and nine closing occurrences were detected across the nine generated dialogues. This high count, which exceeds the number of generated dialogues, suggests greetings were overused.

\begin{figure}[!ht]
\begin{center}
\includegraphics[width=\columnwidth]{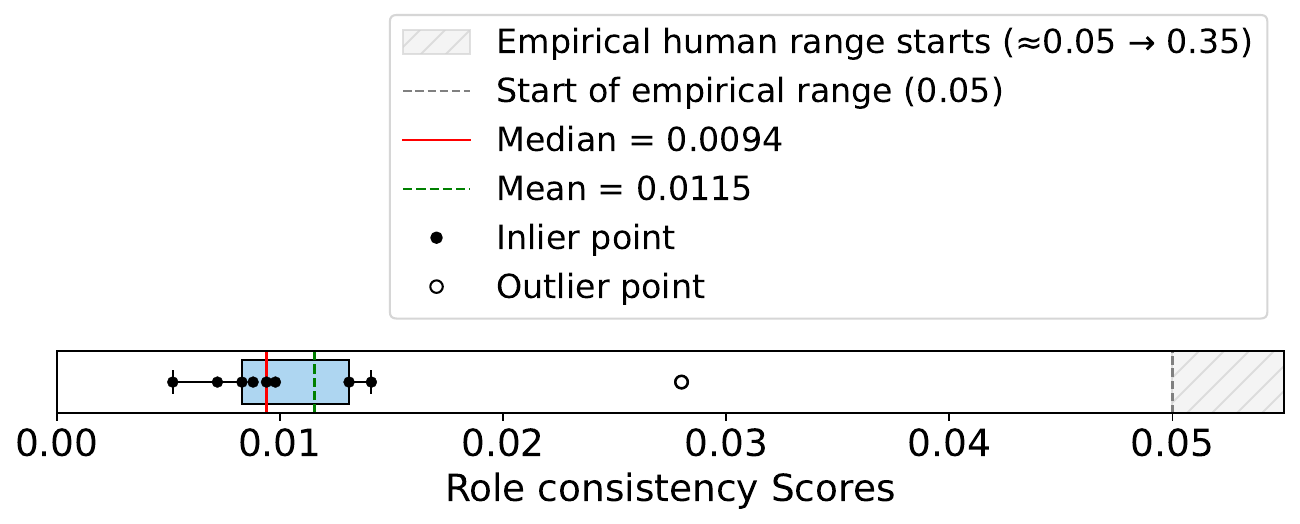}
\caption{Role Consistency - Average per Dialogue}
\label{fig:role_consistency_boxplot}
\end{center}
\end{figure}

\textbf{Role Consistency.}
Keyword matching between each role-specific lexicon and the generated dialogues resulted in a mean overlap score of 0.012. Figure \ref{fig:role_consistency_boxplot} shows a box plot of role consistency scores across the nine dialogues. The clustering of scores within the blue box suggests relatively similar word choice for both doctor and patient across dialogues, with one outlier indicating that one dialogue contained comparatively more role-specific vocabulary. All scores fall below the heuristic baseline reference (gray shaded area).

\begin{figure}[!ht]
\begin{center}
\includegraphics[width=\columnwidth]{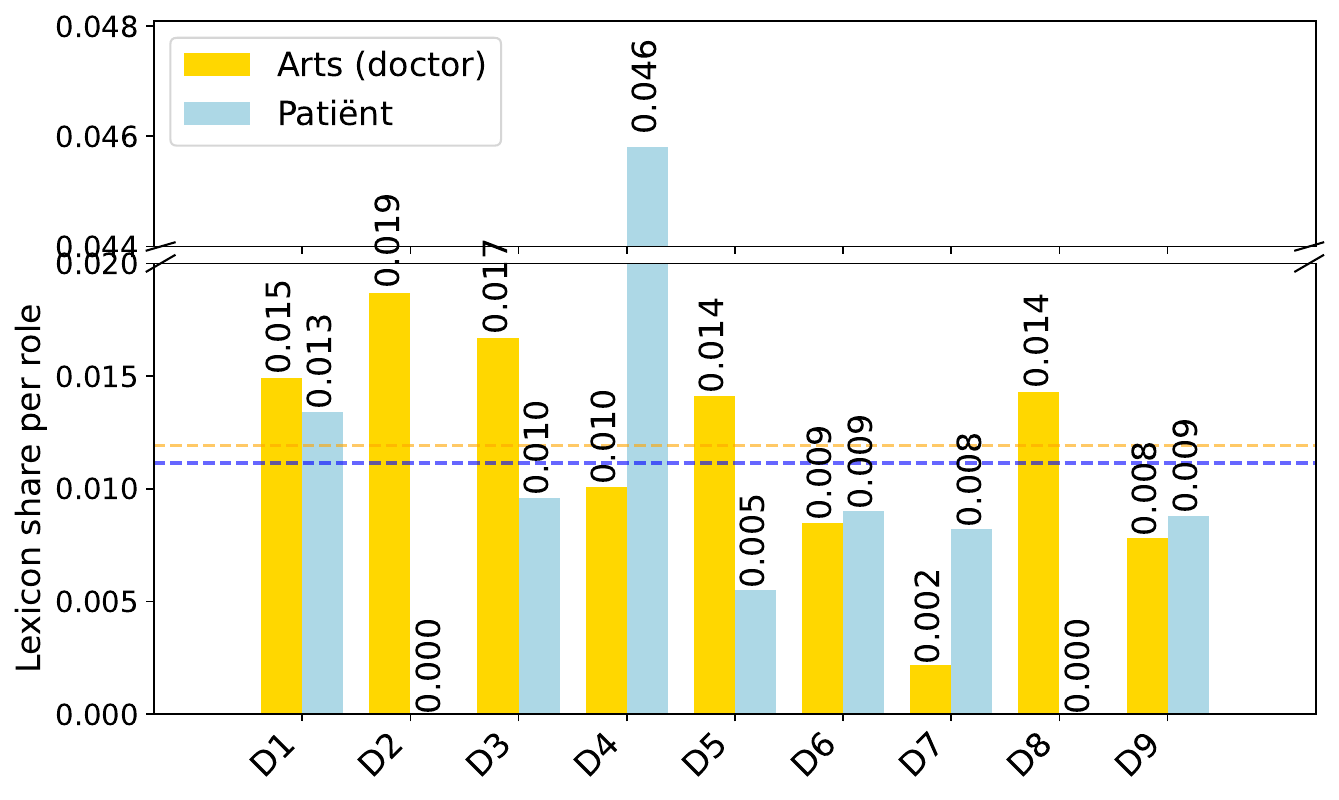}
\caption{Role Consistency - Roles}
\label{fig:role_cutline_bargraph}
\end{center}
\end{figure}

Figure \ref{fig:role_cutline_bargraph} shows the proportion of lexicon match per role within each dialogue. Two dialogues contained no keyword matches for the patient role, while one dialogue showed a higher proportion of patient-specific words.

\begin{figure}[!ht]
\begin{center}
\includegraphics[width=\columnwidth]{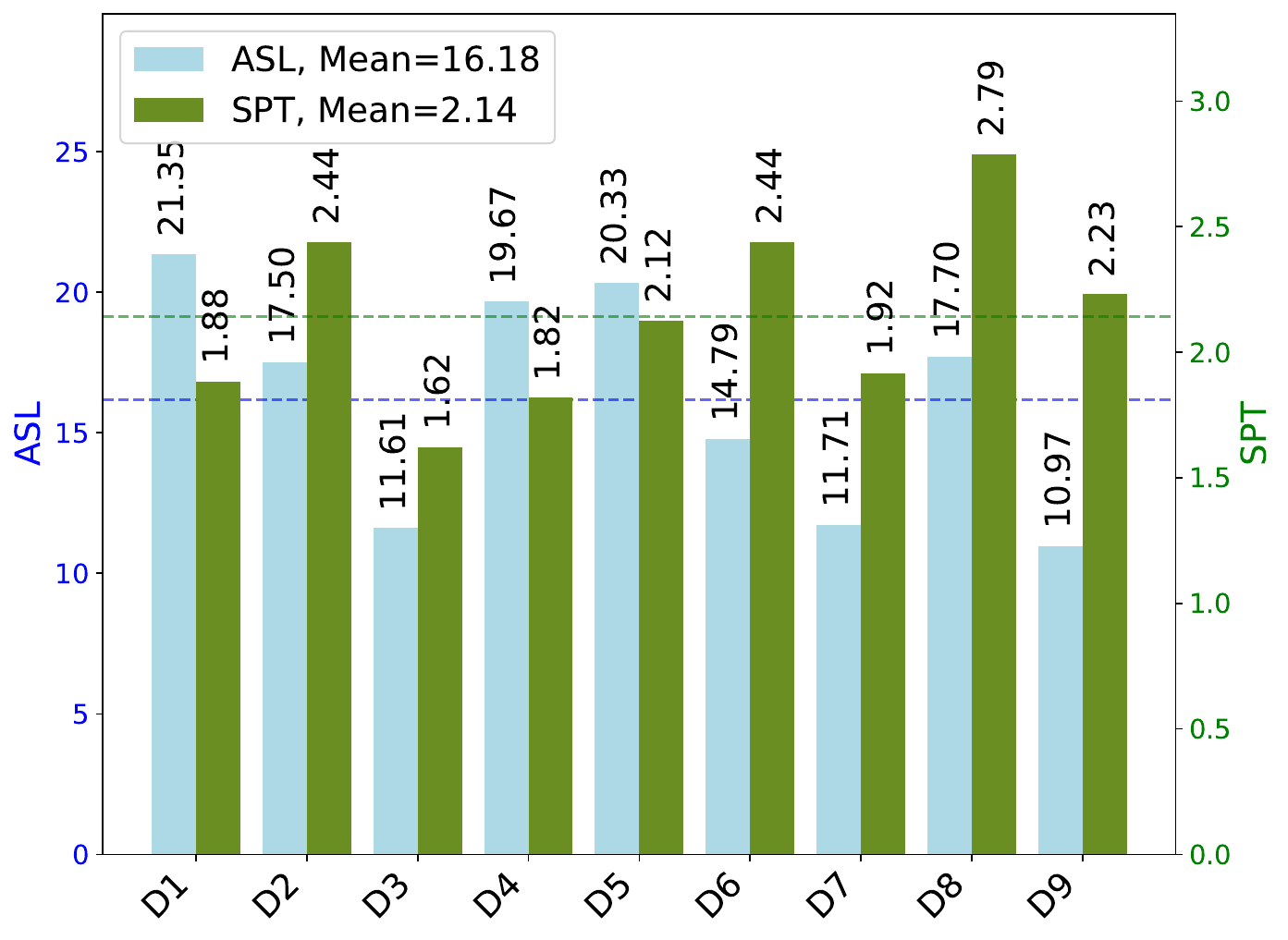}
\caption{Scores: ASL and SPT}
\label{fig:asl_spt}
\end{center}
\end{figure}

\textbf{ASL and Average SPT.}
The ASL of each dialogue ranged between 15 to 20 words, with a mean of 16.18 words, notably higher than the target sample's ASL of seven words. This suggests that the model did not fully adhere to the instruction to produce shorter sentences. The mean SPT was 2.14, with six out of nine dialogues exceeding 1.5, suggesting deviations from the one-sentence-per-turn instruction. Figure \ref{fig:asl_spt} illustrates the ASL and SPT scores across generated dialogues.

\begin{figure}[!ht]
\begin{center}
\includegraphics[width=\columnwidth]{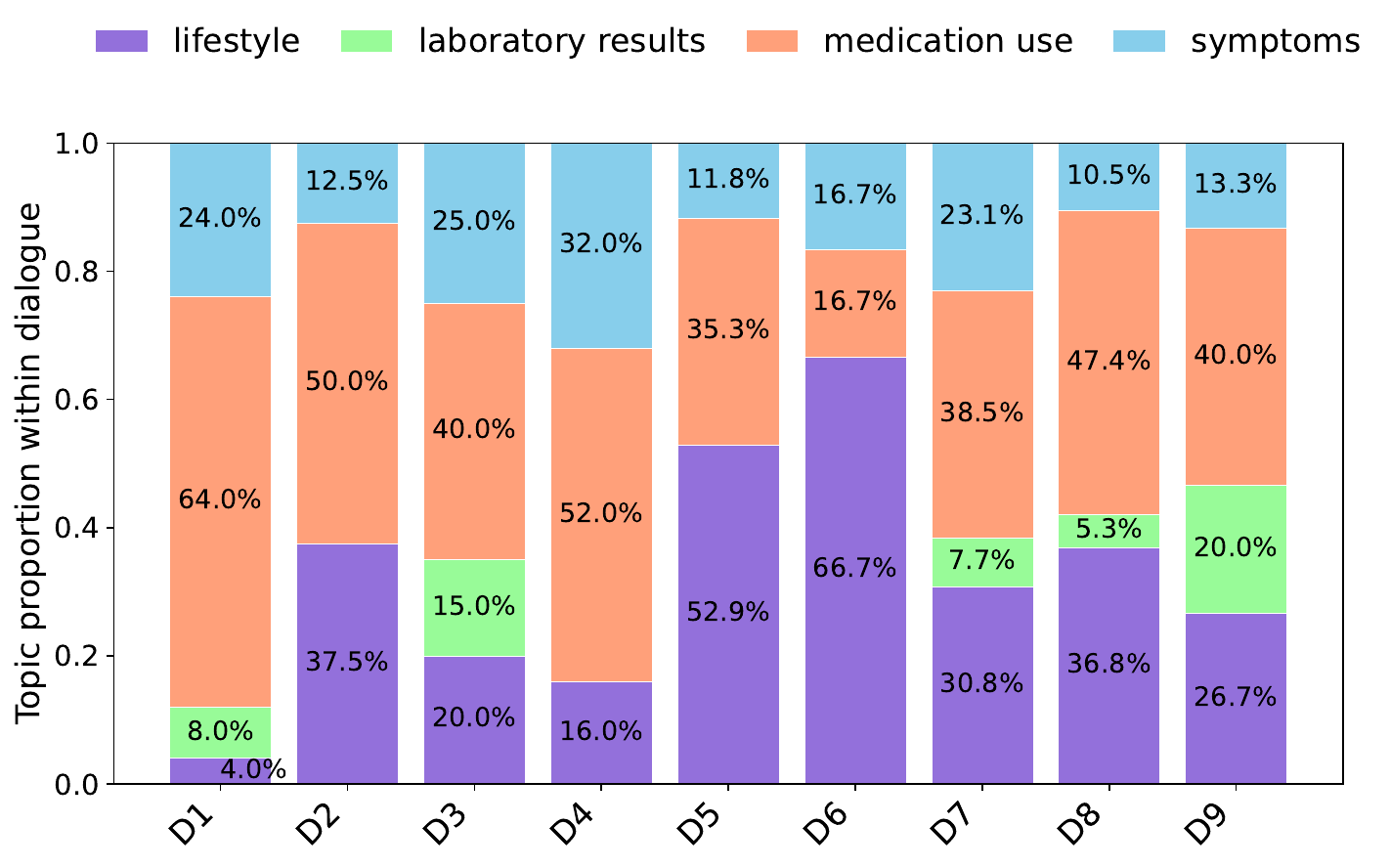}
\caption{Topic Coverage - Proportion per Dialogue}
\label{fig:topic_stacked_graph}
\end{center}
\end{figure}

\textbf{Topic Coverage.}
Keyword matching achieved a mean topic coverage score of 0.889, with reasonable adherence to the prompt instructions. However, topic distribution varied considerably across dialogues, see Figure \ref{fig:topic_stacked_graph}. Discussions on laboratory results were missing in four dialogues, and with minimal coverage in three others, lifestyle and medication use dominated in many dialogues. 

\begin{figure}[!ht]
\begin{center}
\includegraphics[width=\columnwidth]{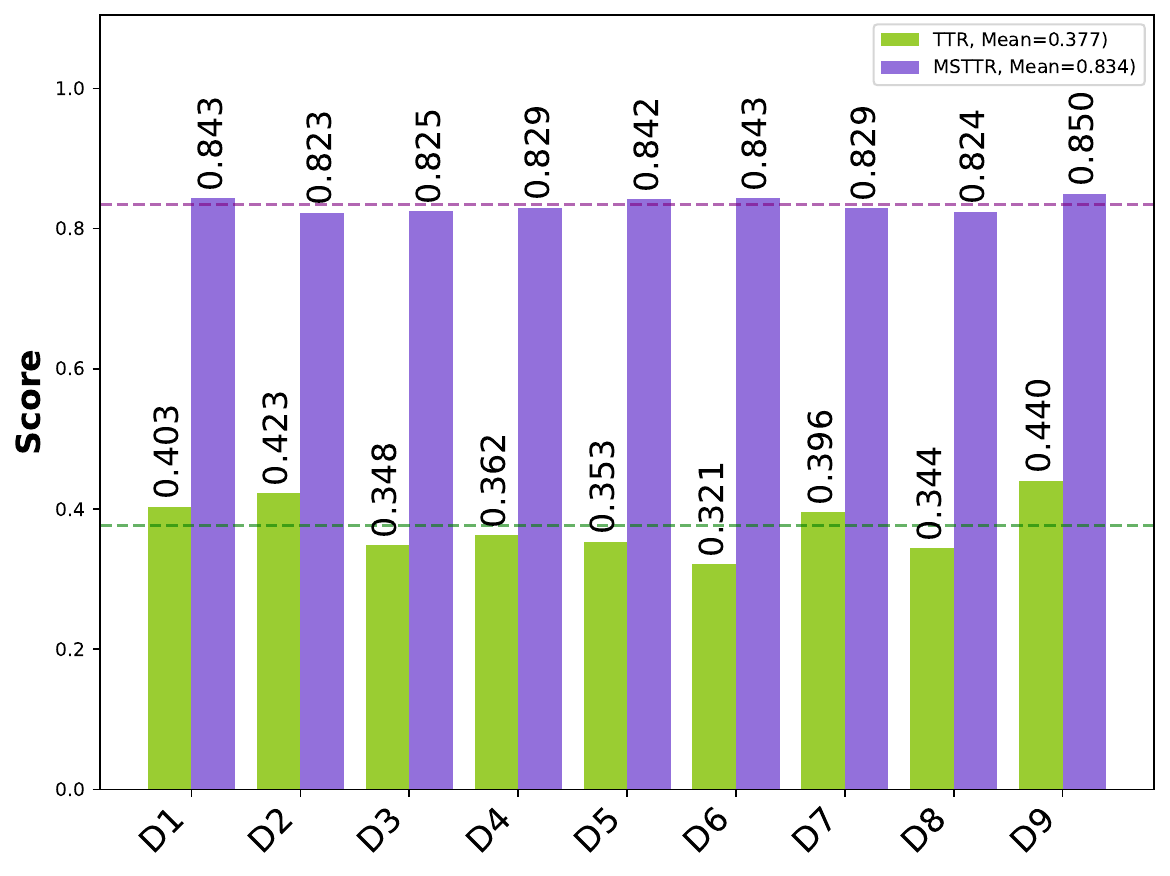}
\caption{Scores: TTR \& MSTTR}
\label{fig:ttr_msttr}
\end{center}
\end{figure}

\textbf{Lexical diversity}
Figure \ref{fig:ttr_msttr} presents the TTR and MSTTR\footnote{\citet{covington2010cutting} proposed the Moving-Average TTR (MATTR), which employs overlapping sliding windows and provides slightly higher precision than MSTTR. In this study, the difference between MSTTR and MATTR was minimal (0.003), and only MSTTR is reported} for all generated dialogues. The scores were largely consistent across dialogues, with a mean TTR = 0.38 and MSTTR = 0.83. The relatively low TTR is expected, given the average dialogue length, as lexical repetition increases with text size, whereas the high MSTTR reflects local lexical variation introduced by specialized medical terminology. 

\subsection{Qualitative Results}
\begin{figure}[!ht]
\begin{center}
\includegraphics[width=\columnwidth]{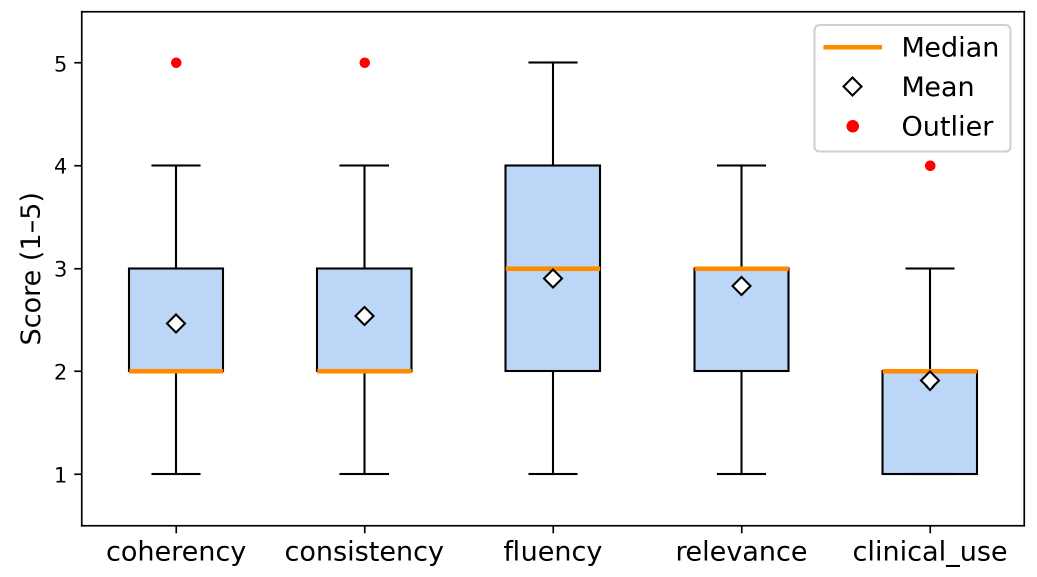}
\caption{Human Evaluation - Scores per Category}
\label{fig:human_cat_avg_box}
\end{center}
\end{figure}

Human raters evaluated the dialogues across five qualitative categories. Figure \ref{fig:human_cat_avg_box} shows the overall score distribution from all raters, with each box representing a category. Scores generally range between two and three, with some exceptions. Coherence and Consistency both exhibit median scores of two, indicating that most ratings cluster around this value. The means for these categories are likely skewed by outliers, represented by red dots in the figure. In contrast, Relevance scores are more tightly concentrated around three, showing that this category received mostly average ratings. Clinical Use scores tend to be low, falling mostly between one and two, with an outlier reflecting a higher score that is not representative of the majority. Fluency received the most varied ratings, with 50\% of data spread between two and four. 

\begin{figure}[!ht]
\begin{center}
\includegraphics[width=\columnwidth]{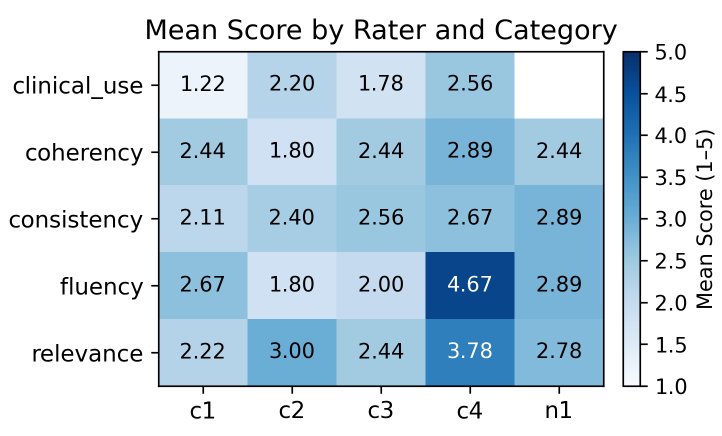}
\caption{Human Evaluation - per Rater in each Category}
\label{fig:human_heatmap}
\end{center}
\end{figure}

The heatmap in Figure \ref{fig:human_heatmap} illustrates the mean score per category by each rater, reflecting potential biases, where C1–C4 denote native Dutch medical practitioners and N1 denotes a native Dutch speaker. Specifically, Rater C4 consistently scored higher for fluency and relevance compared to other raters.

\begin{figure}[!ht]
\begin{center}
\includegraphics[width=\columnwidth]{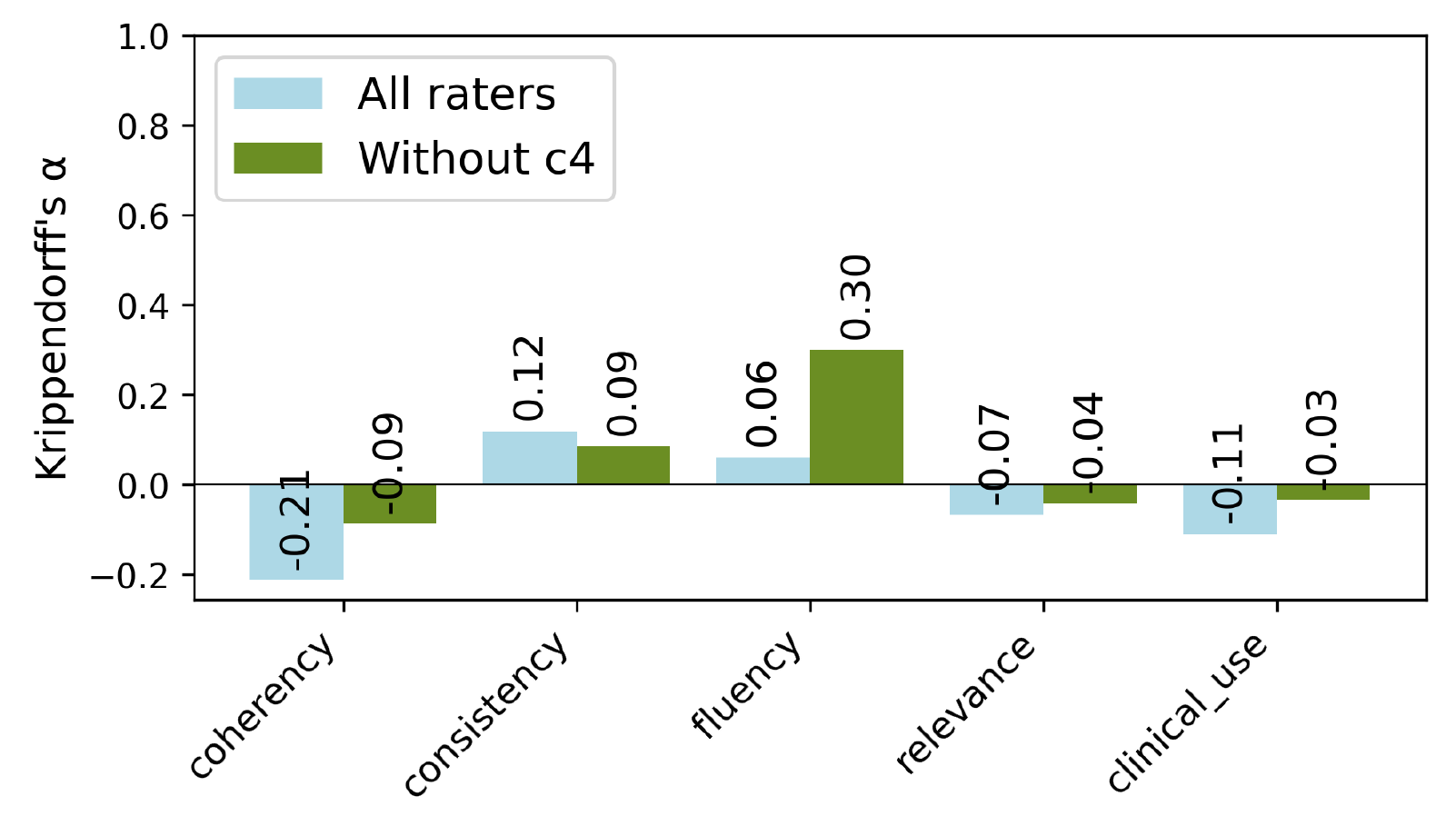}
\caption{Human Evaluation - Inter Rater Reliability}
\label{fig:human_kalpha}
\end{center}
\end{figure}

To further understand rater C4's impact, Figure \ref{fig:human_kalpha} compares IRR measured by Krippendorff's $\alpha$ with and without this rater. Aside from the Fluency category, inclusion of rater C4 had little effect on overall reliability. However, $\alpha$ values remain low across all categories (consistently below 0.12, with 60\% of scores below zero), indicating substantial disagreement among raters. 

\subsection{Comparative Analysis and Discussion}
This section integrates quantitative and qualitative findings to interpret how structural, lexical, and pragmatic aspects jointly influence dialogue quality. Overall, the results reveal that while the model followed the expected conversational structure defined by the prompt, it failed to reflect domain-specific language use and natural variations.

The near-perfect alternation rate suggests highly regular speaker switching, reflecting scripted rather than spontaneous interaction. Closer adherence to the one-sentence-per-turn (SPT) rule would likely lower this value and yield a more natural conversational flow. Frequent greetings, in contrast to the limited number of closings, likely stem from the generation pipeline where dialogues were concatenated from topic-based segments. These effects indicate an over-structured dialogue organization that may be potentially mitigated through prompt refinement or post-processing to smooth transitions. Longer ASL and higher SPT show that the model did not fully adhere to the prompt, representing a separate limitation from the structural overuse seen in alternation rate and greeting use.

Zero keyword matches for some dialogues, low role-specific lexicon overlap, and uneven topic coverage suggest that the model struggles to fully capture context-specific vocabulary. In contrast, uneven topic coverage mirrors real clinical dialogue patterns, where topic shifts do not always follow fixed patterns \citep{robinson2016agenda,ten2002sequential}.

\begin{table}[!ht]
\begin{center}
\footnotesize
\begin{tabularx}{\columnwidth}{lXXX}
\textbf{Measure} & \textbf{D9} & \textbf{D3} & \textbf{D6} \\
\hline
\hline
Length & 610 & 972 & 1266 \\
MSTTR & 0.85 & 0.825 & 0.843 \\
TTR & 0.44 & 0.384 & 0.321 \\
ASL & 10.97 & 11.61 & 14.79 \\
Turns & 26 & 61 & 50 \\
SPT & 2.44 & 1.62 & 2.44 \\

\end{tabularx}
\caption{Comparison of Selected Dialogues by Key Metrics}
\label{tab:quantitative_comparison}
\end{center}
\end{table}

Table \ref{tab:quantitative_comparison} compares three dialogues of varying lengths and their respective scores for MSTTR and ASL. It reveals that dialogue length alone does not determine lexical richness; shorter text achieved comparable MSTTR values to longer ones. It also implies that measured lexical diversity (measured by MSTTR) likely reflects the complexity of medical terminology rather than the role-specific vocabulary use.

It is important to note that lexical validation through context-specific lexicon lists and MSTTR captures only the presence of relevant terms, not their semantic correctness or contextual appropriateness. These measures may also reflect limitations in the lexicon sets or the inherent complexity of Dutch clinical language. Further qualitative analysis is therefore needed to assess contextual accuracy in synthetic dialogues.

Human ratings averaged 2.53, contrasting with high quantitative scores such as alternation rate and lexical diversity measured by MSTTR. Low IRR (Figure \ref{fig:human_kalpha}) likely reflects rubric ambiguity and the subjective nature of evaluation, yet also signals the challenge of capturing conversational naturalness through quantitative measures.

\begin{figure}[!ht]
\begin{center}
\includegraphics[width=\columnwidth]{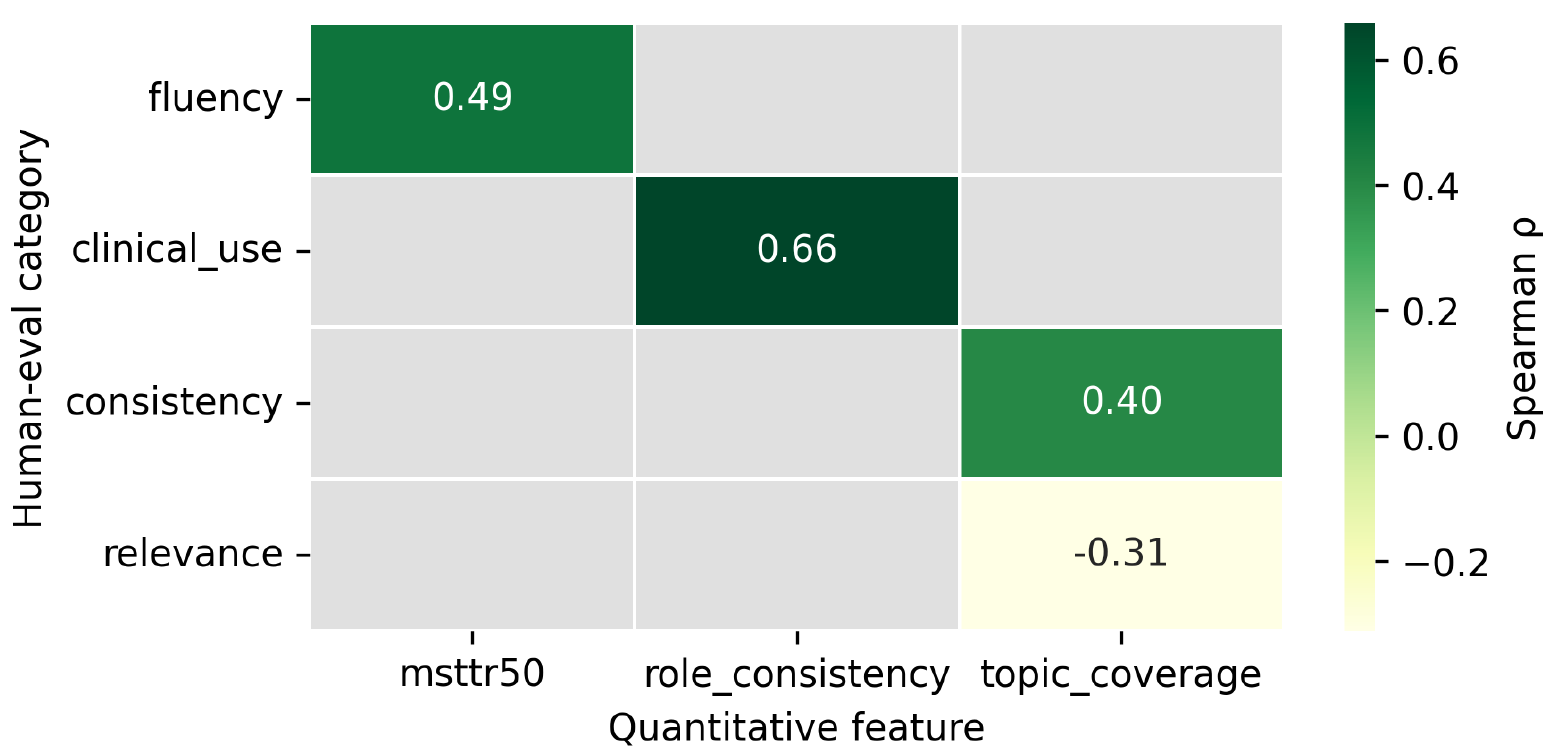}
\caption{Spearman Correlation ($\rho$) - Qualitative vs Quantitative Scores}
\label{fig:spearman_heatmap}
\end{center}
\end{figure}

Correlation analysis (Figure \ref{fig:spearman_heatmap}) confirmed weak alignment between automatic and human assessments: fluency and clinical use correlated moderately with MSTTR and role consistency, whereas relevance showed a negative relation ($\rho$ = –0.31). Given the limited number of dialogues, these correlations should be interpreted with caution. These discrepancies highlight that numeric metrics capture pattern regularity but not semantic or pragmatic naturalness.

These quantitative-qualitative discrepancies echo rater comments, noting issues such as unclear domain focus ("not always clear about the subject being nephrology"), unnatural word choice resembling translated English, and inconsistencies in typical expressions ("errors in typical Flemish expressions"). Additional remarks include multiple greetings, abrupt openings or endings, and unnatural speaker turn-taking patterns different from real clinical conversations. The perception of translation-like output noted by reviewers likely stems from using a model fine-tuned on translated rather than naturally spoken data. Overall, these findings highlight specific weaknesses in the current pipeline - structural over-regularity, lack of domain vocabulary, and limited fluency.

\section{Conclusion and Future Work} 
We proposed a pipeline leveraging a Dutch dataset-fine-tuned LLM to generate synthetic medical dialogues. In answer to our research question, findings indicate that while a Dutch LLM can feasibly produce synthetic medical dialogues that support clinical NLP pipeline development, the generated data do not yet match real-world dialogue quality. Achieving higher quality requires attention to model selection and prompt design, which strongly influence the linguistic features and overall quality of the generated dialogues. Models fine-tuned on translated datasets may negatively affect fluency, whereas overly structured prompts can lead to unnatural or rigid dialogues. These results contribute to ongoing efforts in clinical NLP to mitigate data scarcity and privacy constraints by utilizing realistic, language-specific synthetic datasets.

The main limitations of our study include the lack of domain-specific fine-tuning data, limited computational resources, and constraints on qualitative review process. Future work will focus on improving dialogue generation quality through refined prompt engineering and enhanced human evaluation protocols, including keyword refinement and inter-rater calibration. In addition, downstream development will explore synthetic audio dialogue generation and medical ontology mapping to further improve the realism and clinical usability of the generated dialogues.

Direct comparison with prior work was not possible, as, at the time of submission, no existing studies had addressed synthetic Dutch medical dialogue generation using LLMs. Cross-language comparisons (e.g., with English synthetic dialogue systems) were not pursued, as differences in language and model characteristics would obscure meaningful interpretation.

\section{Ethical Considerations and Limitations}
Synthetic medical dialogues offer an ethically aware alternative in contexts where data scarcity and privacy concerns restrict the development of clinical NLP. Such corpora can be shared in accordance with FAIR principles (Findable, Accessible, Interoperable, Reusable)\footnote{https://www.go-fair.org/fair-principles/}, promoting data sharing and reproducibility without exposing sensitive information.

However, the high similarity observed across generated dialogues suggests that the model may have been adapting to its own synthetic output through iterative testing. Other potential risks include the replication of bias embedded in the training data and potential inaccuracies in clinical content. All dialogues are nephrology-specific, which may limit generalizability to other medical specialties.

Additionally, the few-shot design for dialogue generation prioritized stylistic features over content coherence—input and output segments were drawn from the same file, but likely covered different topics. This may have contributed to high scores on structural metrics (alternation rate, lexical diversity) but lower qualitative ratings for coherence and fluency.

\section{Acknowledgements}
This research was supported by the MediSpeech project funded by ITEA4 under contract number 22032.

We thank qualitative evaluators - Amir Chaman Baz, Lex Dingemans, Sandra van Dulmen, Edwin Geleijn, Henk van den Heuvel - for their comments on synthetic dialogues, which have led to many findings and further improvements.
\section{Bibliographical References}
\bibliographystyle{lrec2026-natbib}
\bibliography{gen_data_reference}


\section{Appendices}

\subsection{Prompt Used For Synthetic Dutch medical Dialogue Generation}\label{app:prompt}
All prompts are written in Dutch.

{\small
\textit{System Prompt, Dutch:}
\begin{quote}
    Je bent een behulpzame medisch onderzoeksassistent die Nederlandstalige medische dialogen genereert. 
    Gebruik alleen 'Patiënt:' en 'Arts:' als sprekerlabels. Gebruik alleen algemeen aanvaarde medische feiten en vermijd het verzinnen van informatie. Corrigeer indien nodig misverstanden in het gesprek. Wees informatief en nauwkeurig binnen het kader van medische kennis. Ga naadloos door zonder gesprekken opnieuw te beginnen of te groeten. Geef geen introductie, vraag geen toestemming, onderbreek niet en produceer alleen dialoog. Geen uitleg, geen voorbeelden, geen meta-commentaar, geen samenvattingen.
\end{quote}

\textit{System Prompt, English translation:}
\begin{quote}
    You are a helpful medical research assistant who generates Dutch medical dialogues.
    Use only 'Patient:' and 'Doctor' as speaker labels. Use only generally accepted medical facts and avoid fabricating information. Correct misunderstandings in the conversation as needed. Be informative and accurate within the framework of medical knowledge. Continue seamlessly without restarting conversations or greeting others. Do not introduce yourself, ask for permission, interrupt, and only generate dialogue. No explanations, no examples, no meta-commentaries, no summaries.
\end{quote}

\textit{User Prompt, Dutch:}
\begin{quote}
    Schrijf een natuurlijke, informele dialoog tussen een patiënt en een art over nefrologie. De dialoog moet minstens 140 gespreksuitingen en ongeveer 1,000 woorden bevatten. Behandel hoofdonderwerpen uit de lijst: symptomen, medicatiegebruik, leefstijl, laboratoriumuitslagen. Gebruik minstens 5 relevante medische vaktermen. Geen opsommingen; integreer lijsten in vraag-en-antwoord. Laat onderwerpen natuurlijk overlopen, gebruik verduidelijkingen en herhalingen voor misverstanden.
    Het gesprek moet vloeiend zijn zonder herhaalde begroetingen of onderbrekingen.  Vervolg het gesprek zonder het opnieuw te starten of scenario's opnieuw te introduceren.
    Speciale kenmerken: veel korte beurten (1–3 woorden), zoals 'ja', 'nee', 'oké', 'hm', afgewisseld met langere, informatieve antwoorden.
    Let op: GEEN uitleg, GEEN voorbeelden, GEEN meta-commentaar, GEEN introductie. Alleen de dialoog.
\end{quote}

\textit{User Prompt, English translation:}
\begin{quote}
    Write a natural, informal dialogue between a patient and a physician about nephrology. The dialogue should contain at least 140 utterances and approximately 1,000 words. Cover main topics from the list: symptoms, medication use, lifestyle, laboratory results. Use at least 5 relevant medical terms. No lists; integrate lists into questions and answers. Allow topics to flow naturally, use Clarifications and repetitions for misunderstandings.
    The conversation should flow smoothly without repeated greetings or interruptions. Continue the conversation without restarting it or reintroducing scenarios.
    Special features: Many short turns (1–3 words), such as 'yes', 'no', 'okay', 'hm', interspersed with longer, informative answers.
    Note: NO explanations, NO examples, NO meta-commentaries, NO introductions. Just the dialogue.
\end{quote}
}

\subsection{Quantitative Evaluation - Keyword Lists}\label{app:keyword-lists}

\textbf{Keyword Lists for Role Consistency Evaluation}
\textit{Role - Doctors, Dutch}
\begin{quote}
    chemotherapie, dialyse, operatie, immunotherapie, therapie, meten, bestralen, vasthouden, transplantatie, injectie, oefenen, hemodialyse, vervangen, voorlichting, vruchtwaterpunctie, bloedtransfusie, euthanasie, roesje, voorschrijven, vervolgonderzoek, fysiotherapie, ondersteuning, screening, discussie, verwijderen, eerste hulp, punctie, PET, conservatieve therapie, beleid, vaccinatie, infusie, voetverzorging, assisteren, bloedtest, evaluatie, voedingsadvies, aanpassing, delegeren, palliatieve zorg, prenatale screening, revisie.
\end{quote}
\textit{Role - Doctors, English translation}
\begin{quote}
    chemotherapy, dialysis, surgery, immunotherapy, therapy, measuring, irradiating, holding, transplantation, injection, exercising, hemodialysis, replacing, information, amniocentesis, blood transfusion, euthanasia, sedation, prescribing, follow-up examination, physiotherapy, support, screening, discussion, removal, first aid, puncture, PET, conservative therapy, policy, vaccination, infusion, foot care, assisting, blood test, evaluation, nutritional advice, adjustment, delegation, palliative care, prenatal screening, revision.
\end{quote}
\textit{Role - Patients, Dutch}
\begin{quote}
    pijn, probleem, ziekte, zwanger, diarree, downsyndroom, hoesten, koorts, speelt, misselijkheid, bevalling, stress, kanker, wil niet, dood, astma, hoest, allergie, geen klachten, bronchitis, trekt, lachen, vermoeidheid, schrijft, drinkt, gevoelig, buikpijn, slaapt, hoofdpijn, slaat, huilen, vloeibaar.
\end{quote}
\textit{Role - Patients, English translation}
\begin{quote}
    pain, problem, illness, pregnant, diarrhea, Down syndrome, coughing, fever, playing, nausea, childbirth, stress, cancer, doesn't want to, death, asthma, cough, allergy, no complaints, bronchitis, pulls, laughs, fatigue, writes, drinks, sensitive, stomach ache, sleeps, headache, hits, cries, fluid.
\end{quote}

\textbf{Keyword Lists for Topic Coverage Evaluation}
\textit{Topic - \textit{symptomen} / symptoms, Dutch}
\begin{quote}
    pijn, hoesten, koorts, misselijkheid, hoest, vermoeidheid, buikpijn, hoofdpijn, spierpijn, keelpijn, vermoeid, duizeligheid, maagpijn, neuropathische pijn, brandende pijn, stekende pijn, aangezichtspijn, abdominale spierpijn, acromioclaviculaire gewrichtspijn.
\end{quote}
\textit{Topic - \textit{symptomen} / symptoms, English translation}
\begin{quote}
    pain, cough, fever, nausea, cough, fatigue, abdominal pain, headache, muscle pain, sore throat, tired, dizziness, stomach pain, neuropathic pain, burning pain, stabbing pain, facial pain, abdominal muscle pain, acromioclavicular joint pain.
\end{quote}
\textit{Topic - \textit{medicatiegebruik} / medication use, Dutch}
\begin{quote}
    medicijnen, kuur, medicijn, medicatie, pil, slikken, paracetamol, recept, dosering, tablet, antibiotica, antibioticum, ibuprofen, toediening, antibiotica-inhalatietherapie, antibiotische therapie, behandelen met bètareceptorantagonist, behandelen met erytropoëtinereceptoragonist, desensitisatiekuur door injectie.
\end{quote}
\textit{Topic - \textit{medicatiegebruik} / medication use, English translation}
\begin{quote}
    medicine, treatment, drug, medication, pill, swallow, paracetamol, prescription, dosage, tablet, antibiotic, antibiotic, ibuprofen, administration, antibiotic inhalation therapy, antibiotic therapy, treatment with beta-receptor antagonist, treatment with erythropoietin receptor agonist, desensitization treatment by injection.
\end{quote}
\textit{Topic - \textit{laboratoriumuitslagen} / laboratory results, Dutch}
\begin{quote}
    glucose, vitamine d, hb, creatinine, cholesterol, bloedtest, kalium, bilirubine, celonderzoek, cervixcytologisch onderzoek, chromosoomonderzoek, crp, d-dimeer, ferritine, ft4, genetisch onderzoek, glucosetolerantietest, hba1c, natrium.
\end{quote}
\textit{Topic - \textit{laboratoriumuitslagen} / laboratory results, English translation}
\begin{quote}
    Glucose, vitamin D, HB, creatinine, cholesterol, blood test, potassium, bilirubin, cell analysis, cervical cytology, chromosome analysis, CRP, D-dimer, ferritin, FT4, genetic testing, glucose tolerance test, HB$\alpha$1C, sodium.
\end{quote}
\textit{Topic - \textit{leefstijl} / lifestyle, Dutch}
\begin{quote}
    suiker, gewicht, slaap, roken, suikerziekte, voeding, stress, dieet, wandelen, inspanning, beweging, zout, rook, oefenen, alcohol, afvallen, sporten, spanning, drank, ontspannen, drankje, afval, fysiek, spannen, aankomen, sport, overgewicht, wijn, bier, voedsel, sportman, voeden, oefening, borrel, sterke drank, lichaamsgewicht, ontspanning, gewichtsverlies, koolhydraat, slaperig.
\end{quote}
\textit{Topic - \textit{leefstijl} / lifestyle, English translation}
\begin{quote}
    sugar, weight, sleep, smoking, diabetes, nutrition, stress, diet, walking, effort, exercise, salt, smoke, practice, alcohol, lose weight, exercise, stress, drink, relax, drink, waste, physical, tense, gain weight, sport, overweight, wine, beer, food, athlete, nourish, exercise, drink, spirits, body weight, relaxation, weight loss, carbohydrate, sleepy.
\end{quote}

\subsection{Qualitative Evaluation Rubric}\label{app:rubric}

\begin{table}[ht]
\begin{center}
\footnotesize
\begin{tabularx}{\columnwidth}{p{0.7cm}X}
\multicolumn{2}{l}{\textbf{Coherency}} \\
\textit{Points} & \textit{Definition and Examples}\\
\hline
\hline
1-2 & Frequent order/timing mistakes (non-sequential turns, confusing switches, inconsistent tense). Dialogue feels unnatural and hard to follow. Ex - doctor and patient turns are frequently mixed up, inserted after goodbye:: "Arts: Goed, laten we nu eens kijken naar wat u al doet en wat we kunnen verbeteren."\\
3 & Mostly sequential and moderately coherent, but some abrupt or unnatural/unclear transitions between topics (e.g. sudden topic change, unclear turn boundaries). Ex - "Patiënt: Ik probeer elke twee uur een half kopje te drinken. Arts: En wat betreft uw voeding?"\\
4-5 & Logically flows, correct turn-taking, consistent tenses, naturally progressing conversation; mostly coherent with minor jumps between topics without clear transition.. Ex - "Patiënt: Ik probeer elke twee uur een half kopje te drinken. Arts: En wat betreft uw voeding?"\\
\end{tabularx}
\end{center}
\end{table}
\begin{table}[ht]
\begin{center}
\footnotesize
\begin{tabularx}{\columnwidth}{p{0.7cm}X}
\multicolumn{2}{l}{\textbf{Consistency}} \\
\textit{Points} & \textit{Definition and Example} \\
\hline
\hline
1-2 & Frequent factual errors, hallucinated medical info, missing or contradictory details with previous statement and context. Lacks accuracy in medical facts, treatment options, or role actions.
Ex - contains hallucinated diseases/treatment - "Arts: Je zegt dat je allergisch bent en Diazepam vijf keer per week neemt." \\
3 & Some inconsistencies, minor errors or missing pieces, e.g., partial topic coverage, minor fact mistakes, not linking blood pressure to kidney function after discussing both.
Ex -  "Patiënt: Ik probeer mijn waterinname te verhogen, maar soms vergeet ik." (General lifestyle is relevant, but lacks specific detail or contradicts earlier plan.)\\
4-5 & Consistent and accurate; mostly or all facts align with scenario and medical knowledge, topics covered.\\
\end{tabularx}
\end{center}
\end{table}
\begin{table}[ht]
\begin{center}
\footnotesize
\begin{tabularx}{\columnwidth}{p{0.7cm}X}
\multicolumn{2}{l}{\textbf{Fluency}} \\
\textit{Points} & \textit{Definition and Example} \\
\hline
\hline
1-2 & Use of Dutch is awkward or ungrammatical, regardless of content. Poor quality sentences, long sentences, repetition in sentences. Translations-like errors.
Ex - "Het rug pijn is niet goed voelen en ik zijn moe altijd is van pijn."\\
3 & Readable with occasional awkwardness, odd phrasing, or grammar slips; generally understandable.
Ex - "Ja, ik ben nemen de pillen en ik voel minder, maar niet blij altijd."\\
4-5 & Fluent, natural, and idiomatic Dutch; well-structured and appropriate medical register.
Ex - "Ik neem ibuprofen en probeer rust te nemen. Maar het helpt niet echt." \\
\end{tabularx}
\end{center}
\end{table}
\begin{table}[ht]
\begin{center}
\footnotesize
\begin{tabularx}{\columnwidth}{p{0.7cm}X}
\multicolumn{2}{l}{\textbf{Relevance}} \\
\textit{Points} & \textit{Definition and Example} \\
\hline
\hline
1-2 & Major topics missing (symptoms, medication use, lifestyle, and laboratory results.), many irrelevant details, or missing clinical actions.
Ex - "Patiënt: Ik houd van voetbal en mijn huisdier is een kat." (Unrelated information to context and/or of topics being discussed.)\\
3 & Some or all target topics touched but detail may be shallow, or topic coverage uneven; some off-topic content. Lacks full coverage. Ex - "Arts: Wat betreft beweging, misschien kunt u een beetje meer wandelen." \\
4-5 & Every target topic is addressed with sufficient detail; all info relevant to consultation.
Ex - "Arts: We kunnen naar fysiotherapie of yoga kijken. Ook uw labresultaten wijzen op een geleidelijk probleem." (Addresses symptoms, medication, lifestyle, and lab results directly.)\\
\end{tabularx}
\end{center}
\end{table}
\begin{table}[t]
\begin{center}
\footnotesize
\begin{tabularx}{\columnwidth}{p{0.7cm}X}
\multicolumn{2}{l}{\textbf{Clinical Use}} \\
\textit{Points} & \textit{Definition and Example} \\
\hline
\hline
1-2 & Dialogue is unrealistic for actual clinical setting (e.g., unsafe advice, fundamental errors, implausible behaviors).
Ex - unsafe or implausible advice is given
ex - "Arts: Neem gewoon meer pijnstillers, zoveel als u wilt, en de rest is niet belangrijk." \\
3 & Could work in clinic with edits; mostly safe and realistic, but has notable weaknesses.
Ex - "Arts: Probeer rustig te blijven en misschien wat meer water drinken." (Safe but incomplete; lacks specific clinical recommendation or next steps.)\\
4-5 & Realistic, safe, plausible for clinical scenario; could be useful for annotation/training.
Ex - Arts: We zullen uw bloeddruk monitoren en indien nodig medicatie voorschrijven. Houd een dagboek bij van uw symptomen en neem contact op als ze verergeren."\\
\end{tabularx}
\end{center}
\end{table}
\end{document}